\documentclass[a4paper,12pt]{article}
\usepackage{amsmath}
\usepackage{amssymb,amsthm,graphicx}
\usepackage{enumitem}
\usepackage{color}
\usepackage{epsfig}
\usepackage{graphics}
\usepackage{pdfpages}
\usepackage{subcaption}
\usepackage[font=small]{caption}
\usepackage[hang,flushmargin]{footmisc} 
\usepackage{float}
\usepackage{rotating,tabularx}
\usepackage{booktabs}
\usepackage[mathscr]{euscript}
\usepackage[numbers]{natbib}
\usepackage{setspace}
\usepackage{mathrsfs}
\usepackage[left=3cm,right=3cm,bottom=3cm,top=3cm]{geometry}
\newcommand{\reals}{\mathbb{R}}

\newcommand{\naturals}{\mathbb{N}}

\newcommand{\ind}{1} 

\newcommand{\bs}[1]{\boldsymbol{#1}} 
\newcommand{\cst}{L}

\theoremstyle{plain}

\newtheorem{theorem}{Theorem}
\newtheorem{lemma}{Lemma}
\newtheorem{corollary}{Corollary}
\newtheorem{definition}{Definition}

\makeatletter
\newcommand{\lefteqno}{\let\veqno\@@leqno}
\makeatother

\newcommand{\heading}[2]
{  \setcounter{page}{1}
   \begin{center}

   \phantom{Distance to upper boundary}
   \vspace{1cm}

   {\LARGE \textbf{#1}} \\[0.4cm]
   {\LARGE \textbf{#2}} 
   \end{center}
}

\newcommand{\authors}[2]
{  \parindent0pt
   \begin{center}
   {\large #1} 
   \vspace{0.1cm}
   
   #2 
   \end{center}
}

\begin{document}

\heading{On the Differences}{between $\boldsymbol{L_2}$Boosting and the Lasso} 
\vspace{-0.2cm}

\authors{Michael Vogt\renewcommand{\thefootnote}{$*$}\footnotemark[1]}{University of Bonn} 
\renewcommand{\thefootnote}{\fnsymbol{footnote}} 
\footnotetext[1]{Address: Department of Economics and Hausdorff Center for Mathematics, University of Bonn, 53113 Bonn, Germany. Email: \texttt{michael.vogt@uni-bonn.de}.}

\vspace{-0.9cm}

\renewcommand{\abstractname}{}
\begin{abstract}
\noindent We prove that $L_2$Boosting lacks a theoretical property which is central to the behaviour of $\ell_1$-penalized methods such as basis pursuit and the Lasso: Whereas $\ell_1$-penalized methods are guaranteed to recover the sparse parameter vector in a high-dimensional linear model under an appropriate restricted nullspace property, $L_2$Boosting is not guaranteed to do so. Hence, $L_2$Boosting behaves quite differently from $\ell_1$-penalized methods when it comes to para\-meter recovery/estimation in high-dimensional linear models. 
\end{abstract}

\renewcommand{\baselinestretch}{1.2}\normalsize
\renewcommand{\thefootnote}{\arabic{footnote}}
\setcounter{footnote}{0}
\numberwithin{equation}{section}
\allowdisplaybreaks[1]

\section{Introduction}

The main aim of this paper is to point out an important theoretical difference between $L_2$Boosting and $\ell_1$-penalized methods such as basis pursuit \cite{ChenDonohoSaunders1998} and the Lasso \cite{Tibshirani1996}. To do so, we consider the high-dimensional linear model 
\begin{equation}\label{model}
\bs{Y} = \bs{X} \bs{\beta}
\end{equation}
without noise, where $\bs{Y} \in \reals^n$ is the observation vector, $\bs{X} \in \reals^{n \times p}$ is the design matrix with $p > n$ and $\bs{\beta} = (\beta_1,\ldots,\beta_p)^\top \in \reals^p$ denotes the parameter vector. Suppose that $\bs{\beta}$ is the unique sparsest solution of the equation $\bs{Y} = \bs{X} \bs{b}$ and let $S = \{ j: \beta_j \ne 0 \}$ be the active set with the sparsity index $s = |S|$. Our main result is negative. It shows that $L_2$Boosting lacks a theoretical property which is central to the behaviour of $\ell_1$-penalized methods such as basis pursuit and the Lasso: Whereas $\ell_1$-penalized methods are guaranteed to recover the sparse parameter vector $\bs{\beta}$ in model \eqref{model} under an appropriate restricted nullspace property, $L_2$Boosting is not guaranteed to do so.

More formally speaking, we prove the following result: Let $\| \cdot \|_1$ be the usual $\ell_1$-norm for vectors and let $S^c = \{1,\ldots,p\} \setminus S$ be the complement of $S$. Moreover, for any vector $\bs{b} = (b_1,\ldots,b_p)^\top \in \reals^p$ and any index set $\mathcal{T} \subseteq \{1,\ldots,p\}$, define $\bs{b}_{\mathcal{T}} = (b_{\mathcal{T},1},\ldots,b_{\mathcal{T},p})^\top \in \reals^p$ by setting $b_{\mathcal{T},j} = b_j \, \ind(j \in \mathcal{T})$ for $1 \le j \le p$. The design matrix $\bs{X}$ is said to fulfill the restricted nullspace property $\text{RN}(S,\cst)$ for the index set $S$ and the constant $\cst > 0$ if the cone $\mathbb{C}(S,\cst) = \{ \bs{b} \in \reals^p: \| \bs{b}_{S^c} \|_1 \le \cst \| \bs{b}_{S} \|_1 \}$ and the nullspace $\mathcal{N}(\bs{X})$ of $\bs{X}$ only have the zero vector in common, that is, if $\mathbb{C}(S,\cst) \, \cap \, \mathcal{N}(\bs{X}) = \{ \bs{0} \}$. As is well-known, basis pursuit and the Lasso are guaranteed to recover the sparse vector $\bs{\beta}$ under the restricted nullspace property $\text{RN}(S,\cst)$ with $\cst \ge 1$. We prove that $L_2$Boosting, in contrast, may fail to recover $\bs{\beta}$ under $\text{RN}(S,\cst)$ no matter how large $\cst$. In particular, for any $\cst > 0$, we construct a matrix $\bs{X}$ with the property $\text{RN}(S,\cst)$ and a vector $\bs{\beta}$ which is the unique sparsest solution of $\bs{Y} = \bs{X} \bs{b}$ such that the parameter estimate $\bs{\beta}^{[k]}$ produced by the $L_2$Boosting algorithm in the $k$-th iteration step does not converge to $\bs{\beta}$ as $k \rightarrow \infty$. Hence, $L_2$Boosting fails to recover the sparse parameter vector $\bs{\beta}$.

According to this negative result (which, to the best of our knowledge, has not been known so far), $L_2$Boosting behaves quite differently from $\ell_1$-penalized methods when it comes to parameter recovery/estimation in high-dimensional linear models. This comes a bit as a surprise. As $L_2$Boosting is usually considered to be closely related to $\ell_1$-penalized methods, one may have expected that it is guaranteed to recover the parameter vector $\bs{\beta}$ under $\text{RN}(S,\cst)$ at least for sufficiently large $\cst$. There are indeed important connections between $L_2$Boosting and $\ell_1$-penalized methods such as the Lasso. In a very influential paper, Efron et al.\ \cite{Efronetal2004} established close similarities between the Lasso and forward stagewise linear regression (FSLR), which is a near relative of $L_2$Boosting. They proved that FSLR (with step size approaching zero) and the Lasso can be obtained as slight modifications of the least angle regression (LARS) algorithm. In addition, they showed that FSLR (with step size approaching zero) and the Lasso coincide under a positive cone condition, which is related to but more general than orthogonality of the design matrix. Exact equivalence of $L_2$Boosting and the Lasso in an orthonormal linear model was proven in B\"uhlmann \& Yu \cite{BuehlmannYu2006}. Despite these close connections, $L_2$Boosting and the Lasso are not the same in general but may produce different solutions. This has been observed in numerous simulation and application studies; cp.\ for example Hepp et al.\ \cite{Heppetal2016} for a recent review. The exact theoretical reasons why $L_2$Boosting and the Lasso may produce quite different solutions are however not fully understood so far. The negative result of this paper contributes to better understand these reasons.

\section{Notation and definitions}

\subsection{Notation}

We briefly summarize the notation used in the paper. As already mentioned in the Introduction, for a general index set $\mathcal{T} \subseteq \{1,\ldots,p\}$ and any vector $\bs{b} = (b_1,\ldots,b_p)^\top \in \reals^p$, we define the vector $\bs{b}_{\mathcal{T}} = (b_{\mathcal{T},1},\ldots,b_{\mathcal{T},p})^\top \in \reals^p$ by setting $b_{\mathcal{T},j} = b_j \, \ind(j \in \mathcal{T})$ for $1 \le j \le p$. We further write $\bs{X} = (\bs{X}_1,\ldots,\bs{X}_p)$, where $\bs{X}_j$ denotes the $j$-th column of the design matrix $\bs{X}$. As usual, the symbol $\| v \|_q$ denotes the $\ell_q$-norm of a generic vector $v \in \reals^N$ for $q \in \naturals \cup \{ \infty \}$. In addition, we use the symbol $\langle v, w \rangle = v^\top w$ to denote the inner product of vectors $v, w \in \reals^N$. Finally, the cardinality of a set $\mathcal{T}$ is denoted by $| \mathcal{T} |$.

\subsection{$\bs{L_2}$Boosting}

Boosting methods were originally proposed for classification and go back to Schapire \cite{Schapire1990} and Freund \& Schapire \cite{FreundSchapire1996}. Since then, a variety of boosting algorithms have been developed for different purposes. The $L_2$Boosting algorithm has been investigated in the statistics, the signal processing and the approximation literature under different names. In statistics, $L_2$Boosting methods for regression were developed by Friedman \cite{Friedman2001}. In signal processing, $L_2$Boosting is known as matching pursuit and was introduced by Mallat \& Zhang \cite{MallatZhang1993} and Qian \& Chen \cite{QianChen1994}. In the approximation literature, it goes under the name of pure greedy algorithm; cp.\ for example Temlyakov's monograph \cite{Temlyakov2011}. Prediction consistency of $L_2$Boosting in high-dimensional linear models was derived by Temlyakov \cite{Temlyakov2000} in the noiseless case and by B\"uhlmann \cite{Buehlmann2006} in the noisy case. Results on support recovery for greedy algorithms related to $L_2$Boosting were established in Tropp \cite{Tropp2004} and Donoho et al.\ \cite{DonohoEladTemlyakov2006} among others.

The $L_2$Boosting algorithm proceeds as follows: 
\begin{itemize}[leftmargin=1.5cm]
\item[Step 0:] Initialize the residual and parameter vector by $\bs{R}^{[0]} = \bs{Y}$ and $\bs{\beta}^{[0]} = \bs{0} \in \reals^p$.  
\item[Step $k$:] Let $\bs{R}^{[k-1]} \in \reals^n$ and $\bs{\beta}^{[k-1]} \in \reals^p$ be the residual and parameter vector from the previous iteration step. For each $1 \le j \le p$, compute the univariate least-squares estimate $\hat{b}_j = \langle \bs{R}^{[k-1]}, \bs{X}_j \rangle / \| \bs{X}_j \|_2^2$ and define the index  $j_k \in \{1,\ldots,p\}$ by
\begin{equation}\label{alg-index1}
j_k = \arg\min_{1 \le j \le p} \big\| \bs{R}^{[k-1]} - \hat{b}_j \bs{X}_j \big\|_2^2. 
\end{equation}
Equivalently, $j_k$ can be defined as 
\begin{equation}\label{alg-index2}
j_k = \arg\max_{1 \le j \le p} \Big| \big\langle \bs{R}^{[k-1]}, \frac{\bs{X}_{j}}{\| \bs{X}_{j} \|_2} \big\rangle \Big|. 
\end{equation}
In case of ties, let $j_k$ be the smallest index which fulfills \eqref{alg-index1}. Update $\bs{R}^{[k-1]}$ and $\bs{\beta}^{[k-1]}$ by 
\begin{equation}
\bs{R}^{[k]} = \bs{R}^{[k-1]} - \nu \hat{b}_{j_k} \bs{X}_{j_k}
\end{equation}
and   
\begin{equation}
\bs{\beta}^{[k]} = \bs{\beta}^{[k-1]} + \nu \hat{b}_{j_k} \bs{e}_{j_k},
\end{equation}
where $\bs{e}_j$ is the $j$-th standard basis vector of $\reals^p$ and $\nu \in (0,1]$ is a pre-specified step length. 
\end{itemize}
Iterate this procedure until some stopping criterion is satisfied or until some maximal number of iterations is reached.

\subsection{Restricted nullspace, restricted eigenvalue and restricted isometry properties}

Let $\mathcal{T} \subseteq \{1,\ldots,p\}$ be an arbitrary index set and let $\cst$ be a positive real constant. Define the cone $\mathbb{C}(\mathcal{T},\cst) = \{ \bs{b} \in \reals^p: \| \bs{b}_{\mathcal{T}^c} \|_1 \le \cst \| \bs{b}_{\mathcal{T}} \|_1 \}$ and denote the nullspace of $\bs{X}$ by $\mathcal{N}(\bs{X})$.
\begin{definition}
The design matrix $\bs{X} \in \reals^{n \times p}$ satisfies the restricted nullspace property $\textnormal{RN}(\mathcal{T},\cst)$ for the index set $\mathcal{T}$ and the constant $\cst$ if
\[ \mathbb{C}(\mathcal{T},\cst) \cap \mathcal{N}(\bs{X}) = \{ \bs{0} \}. \]
If $\bs{X}$ satisfies $\textnormal{RN}(\mathcal{T},\cst)$ for any index set $\mathcal{T}$ with $|\mathcal{T}| \le t$, we say that it fulfills the uniform restricted nullspace property $\textnormal{RN}_{\textnormal{unif}}(t,\cst)$ of order $t$. 
\end{definition}
The $\text{RN}(\mathcal{T},\cst)$ property restricts the nullspace $\mathcal{N}(\bs{X})$ not to intersect with the cone $\mathbb{C}(\mathcal{T},\cst)$ except at zero. The cone region $\mathbb{C}(\mathcal{T},\cst)$ gets larger with increasing $\cst$. Hence, the $\text{RN}(\mathcal{T},\cst)$ property becomes more restrictive as $\cst$ increases. Further discussion of the restricted nullspace property can be found in Donoho \& Huo \cite{DonohoHuo2001}, Feuer \& Nemirovski \cite{FeuerNemirovski2003} and Cohen, Dahmen \& DeVore \cite{CohenDahmenDeVore2009} among others.

The $\text{RN}(\mathcal{T},\cst)$ property is closely related to restricted eigenvalue properties which were introduced in Bickel et al.\ \cite{BickelRitovTsybakov2009} and are frequently used in the context of the Lasso. In particular, $\text{RN}(\mathcal{T},\cst)$ is equivalent to the following restricted eigenvalue property. 
\begin{definition}
The design matrix $\bs{X} \in \reals^{n \times p}$ satisfies the restricted eigenvalue property $\textnormal{RE}(\mathcal{T},\cst)$ for the index set $\mathcal{T}$ and the constant $\cst$ if there exists a constant $\phi > 0$ with
\[ \frac{\| \bs{X} \bs{b} \|_2^2}{\| \bs{b} \|_2^2} \ge \phi \quad \text{for all non-zero } \bs{b} \in \mathbb{C}(\mathcal{T},\cst). \]
\end{definition}
Sufficient conditions for restricted nullspace and eigenvalue properties are often formulated in terms of restricted isometry \cite{CandesTao2005}. The $t$-restricted isometry constant $\delta_t$ of the matrix $\bs{X}$ is defined as the smallest non-negative number such that 
\[ (1 - \delta_t) \le \frac{\| \bs{X} \bs{b} \|_2^2}{\| \bs{b} \|_2^2} \le (1 + \delta_t) \] 
for any non-zero $\mathcal{T}$-sparse vector $\bs{b}$ whose active set $\mathcal{T}$ has cardinality $|\mathcal{T}| \le t$. There are several results in the literature which show that the uniform restricted nullspace property $\textnormal{RN}_{\textnormal{unif}}(t,1)$ holds true if the restricted isometry constants $\delta_t$ and $\delta_{2t}$ fulfill certain conditions. An example is the following result: If the restricted isometry constant $\delta_{2t}$ of the matrix $\bs{X}$ is such that $\delta_{2t} < 1/3$, then $\bs{X}$ has the uniform restricted nullspace property $\textnormal{RN}_{\textnormal{unif}}(t,1)$. Conceptually, restricted isometry conditions are substantially stronger than restricted nullspace/eigenvalue conditions: Restricted isometry conditions require that both a lower and upper bound of the form $(1-\delta) \le \| \bs{X} \bs{b} \|_2^2 / \| \bs{b} \|_2^2 \le (1+\delta)$ hold for all vectors $\bs{b}$ that fulfill certain sparsity constraints. Restricted nullspace/eigenvalue conditions, in contrast, only require a lower bound of the form $(1-\delta) \le \| \bs{X} \bs{b} \|_2^2 / \| \bs{b} \|_2^2$ to hold for all vectors $\bs{b}$ with certain sparsity properties.

\section{The main result}\label{sec-result}

Consider the high-dimensional linear model $\bs{Y} = \bs{X} \bs{\beta}$ without noise from \eqref{model}, where $\bs{X} \in \reals^{n \times p}$ is the design matrix and $\bs{\beta} \in \reals^p$ is the sparsest possible parameter vector. As before, we denote the active set by $S = \{ j: \beta_j \ne 0 \}$ and its cardinality by $s = |S|$. According to the following theorem, $L_2$Boosting may fail to recover the vector $\bs{\beta}$ under the uniform restricted nullspace property $\textnormal{RN}_{\textnormal{unif}}(s,\cst)$ no matter how large $\cst$. 
\begin{theorem}\label{theo-counterexample}
For any $\cst > 0$, there exist 
\begin{enumerate}[label=(\alph*),leftmargin=0.75cm]
\item \label{cond-a-theo} a design matrix $\bs{X} \in \reals^{n \times p}$ for some $n$ and $p$ with $n < p$ which fulfills the uniform restricted nullspace property $\textnormal{RN}_{\textnormal{unif}}(s,\cst)$ and
\item \label{cond-b-theo} a vector $\bs{\beta} \in \reals^p$ with $|S| = s$ which is the unique sparsest solution of the equation $\bs{Y} = \bs{X} \bs{b}$
\end{enumerate}
such that 
\[ \| \bs{\beta}^{[k]} - \bs{\beta} \|_1 \not\rightarrow 0 \quad \text{as } k \rightarrow \infty. \]
Hence, the $S$-sparse vector $\bs{\beta}$ is not recovered by $L_2$Boosting. 
\end{theorem}
The proof of Theorem \ref{theo-counterexample} is given in Section \ref{subsec-result-proof-theo}. There, we construct a design matrix $\bs{X}$ and a parameter vector $\bs{\beta}$ for each $\cst > 0$ which satisfy conditions \ref{cond-a-theo} and \ref{cond-b-theo} and for which the following holds: For any $k \ge 0$, the parameter vector $\bs{\beta}^{[k]} = \bs{\beta}_S^{[k]} + \bs{\beta}_{S^c}^{[k]}$ produced by the boosting algorithm in the $k$-th iteration step is such that $\bs{\beta}_S^{[k]} = \bs{0}$. Hence, $L_2$Boosting never selects an index $j$ in the active set $S$. This implies that $\| \bs{\beta}^{[k]} - \bs{\beta} \|_1 \ge \| \bs{\beta}_S \|_1$ for any $k$, which in turn yields that $\| \bs{\beta}^{[k]} - \bs{\beta} \|_1 \not\rightarrow 0$ as $k \rightarrow \infty$.

\subsection{Discussion of Theorem \ref{theo-counterexample}}\label{subsec-result-discussion}

In what follows, we compare the behaviour of $L_2$Boosting specified in Theorem \ref{theo-counterexample} with that of $\ell_1$-penalized methods such as basis pursuit and the Lasso. Similar points apply to the Dantzig selector of Cand{\`e}s \& Tao \cite{CandesTao2007}. For brevity, we however restrict attention to basis pursuit and the Lasso. Basis pursuit approximates the sparse parameter vector $\bs{\beta}$ in model \eqref{model} by any solution $\bs{\beta}^{\text{BP}}$ of the minimization problem 
\begin{equation}\label{def-BP}
\underset{\bs{b} \in \reals^p}{\text{minimize}} \, \| \bs{b} \|_1 \quad \text{subject to } \bs{Y} = \bs{X} \bs{b},
\end{equation}
whereas the Lasso (in its Lagrangian form) is defined as any solution $\bs{\beta}_\lambda^{\text{Lasso}}$ of the problem 
\begin{equation}\label{def-lasso}
\underset{\bs{b} \in \reals^p}{\text{minimize}} \, \big\{ \| \bs{Y} - \bs{X} \bs{b} \|_2^2 + \lambda \| \bs{b} \|_1 \big\} 
\end{equation}
with $\lambda > 0$ denoting the penalty constant.

In contrast to $L_2$Boosting, basis pursuit and the Lasso are guaranteed to recover the $S$-sparse vector $\bs{\beta}$ in the high-dimensional linear model \eqref{model} under an appropriate restricted nullspace property. More precisely, if the design matrix $\bs{X}$ fulfills $\text{RN}(S,\cst)$ with $\cst \ge 1$, then $\bs{\beta}$ is the unique solution of the minimization problem \eqref{def-BP}, that is, $\bs{\beta} = \bs{\beta}^{\text{BP}}$. Moreover, under $\text{RN}(S,\cst)$ with $\cst \ge 1$, it holds that $\bs{\beta} = \lim_{\lambda \rightarrow 0} \bs{\beta}_\lambda^{\text{Lasso}}$. Hence, $\bs{\beta}$ is recovered as the limit of the Lasso estimator $\bs{\beta}_\lambda^{\text{Lasso}}$, where the penalty constant $\lambda$ converges to zero. Letting the penalty constant $\lambda$ of the Lasso estimator converge to zero corresponds to letting the number of boosting iterations $k$ go to infinity.

The main reason why basis pursuit and the Lasso are ensured to recover the vector $\bs{\beta}$ under the nullspace contraint $\text{RN}(S,\cst)$ with $\cst \ge 1$ is the following: The residuals $\Delta^{\text{BS}} = \bs{\beta}^{\text{BS}} - \bs{\beta}$ and $\Delta^{\text{Lasso}}_\lambda = \bs{\beta}^{\text{Lasso}}_\lambda - \bs{\beta}$ are guaranteed to lie in the cone $\mathbb{C}(S,1)$, that is, 
\[ \| \Delta^{\text{BS}}_{S^c} \|_1 \le \| \Delta^{\text{BS}}_{S} \|_1 \quad \text{and} \quad \| \Delta^{\text{Lasso}}_{\lambda,S^c} \|_1 \le \| \Delta^{\text{Lasso}}_{\lambda,S} \|_1 \quad \text{for any } \lambda > 0. \]
$L_2$Boosting, in contrast, does not have this property. In particular, we can show the following result on the boosting residuals $\Delta^{[k]} = \bs{\beta}^{[k]} - \bs{\beta}$. 
\begin{corollary}\label{corollary-counterexample} 
For any $\cst > 0$, there exist a design matrix $\bs{X}$ and a para\-meter vector $\bs{\beta}$ with the properties \ref{cond-a-theo} and \ref{cond-b-theo} from Theorem \ref{theo-counterexample} such that 
\[ \Delta^{[k]} \notin \mathbb{C}(S,\cst) \quad \text{for sufficiently large } k. \] 
\end{corollary} 
The proof of Corollary \ref{corollary-counterexample} is provided in Section \ref{subsec-result-proof-corollary}.

\subsection{Proof of Theorem \ref{theo-counterexample}}\label{subsec-result-proof-theo}

For any given constant $\cst > 0$, we consider the following design: We let $p = n+1$ and choose $n$ to be a natural number whose square root is a natural number itself, that is, $n = N^2$ for some $N \in \naturals$. We pick $n$ sufficiently large, in particular so large that 
\begin{equation*}
n \ge 5 \quad \text{and} \quad \frac{n+1-\sqrt{n}}{\sqrt{n}} > \cst. 
\end{equation*}
The design matrix is given by 
\begin{equation*}
\bs{X} = \begin{array}{c@{\!\!\!}l}
\left( \begin{array}{ccc|ccc|c}
\gamma  &        &         &        &        &        & \gamma  \\
        & \ddots &         &        &        &        & \vdots  \\
        &        & \gamma  &        &        &        & \gamma  \\
\hline
        &        &         & 1      &        &        & 1       \\
        &        &         &        & \ddots &        & \vdots  \\
        &        &         &        &        & 1      & 1       \\
\end{array} \right)
&
\begin{array}[c]{@{}l@{\,}l}
\left. \begin{array}{c} \vphantom{0} \\ \vphantom{\vdots} \\ \vphantom{0} \end{array} \right\} & \text{$s$ times} \\
\left. \begin{array}{c} \vphantom{0} \\ \vphantom{\vdots} \\ \vphantom{0}  \end{array} \right\} & \text{$n-s$ times} \\
\end{array}
\end{array}
\end{equation*}
with $s = \sqrt{n}$ and $\gamma = n$, where empty positions in the matrix correspond to the entry $0$.\footnote{Note that other choices of $s$ and $\gamma$ are possible.} The parameter vector is chosen as 
\[ \bs{\beta} = (\underbrace{1,\ldots,1}_{\text{$s$ times}}, \underbrace{0,\ldots,0}_{\text{$p-s$ times}})^\top \in \reals^p. \]
Hence, the active set is $S = \{1,\ldots,s\}$ and the observation vector $\bs{Y}$ is given by    
\[ \bs{Y} = (\underbrace{\gamma,\ldots,\gamma}_{\text{$s$ times}}, \underbrace{0,\ldots,0}_{\text{$n-s$ times}})^\top \in \reals^n. \]

We first show that conditions \ref{cond-a-theo} and \ref{cond-b-theo} are fulfilled for our choices of $\bs{X}$ and $\bs{\beta}$: The nullspace $\mathcal{N}(\bs{X})$ of $\bs{X}$ is one-dimensional and spanned by the vector $\bs{z} = (-1,\ldots,-1,1) \in \reals^p$. For any subset $\mathcal{T} \subseteq \{1,\ldots,p\}$ with $|\mathcal{T}| \le s$, it holds that  
$\| \bs{z}_\mathcal{T} \|_1 \le s = \sqrt{n}$ and $\| \bs{z}_{\mathcal{T}^c} \|_1 \ge p-s = n+1-\sqrt{n}$, which implies that 
\[ \frac{\| \bs{z}_\mathcal{T} \|_1}{\| \bs{z}_{\mathcal{T}^c} \|_1} \le \frac{\sqrt{n}}{n+1-\sqrt{n}} \rightarrow 0 \quad \text{as } n \rightarrow \infty. \]
From this, it immediately follows that the design matrix $\bs{X}$ satisfies the uniform restricted nullspace property $\text{RN}_{\text{unif}}(s,\cst)$ for any
\begin{equation*}
\cst < \frac{n+1-\sqrt{n}}{\sqrt{n}}. 
\end{equation*}
Hence, condition \ref{cond-a-theo} is fulfilled. In order to see that condition (b) is satisfied as well, we make use of the following fact which is a consequence of results due to Donoho \& Elad \cite{DonohoElad2003} and Gribonval \& Nielsen \cite{GribonvalNielsen2003}: The vector $\bs{\beta}$ is the unique sparsest solution of the equation $\bs{Y} = \bs{X} \bs{b}$ if $s < \text{spark}(\bs{X})/2$, where $\text{spark}(\bs{X})$ is the least number of columns of $\bs{X}$ that form a linearly dependent set. Since $s = \sqrt{n}$ and $\text{spark}(\bs{X}) = n$, the inequality $s < \text{spark}(\bs{X})/2$ holds for any $n \ge 5$. Consequently, the parameter vector $\bs{\beta}$ is guaranteed to satisfy condition \ref{cond-b-theo} for any $n \ge 5$.

We now prove that $\| \bs{\beta}^{[k]} - \bs{\beta} \|_1 \not\rightarrow 0$ as $k \rightarrow \infty$. To do so, we verify the following lemma.
\begin{lemma}\label{lemma-counterexample} 
Suppose that the vector $\bs{\beta}^{[k]}$ obtained in the $k$-th iteration step of the boosting algorithm has the form 
\begin{equation}\label{beta-k}
\bs{\beta}^{[k]} = (\underbrace{0,\ldots,0}_{\text{$s$ \textnormal{times}}}, -c_{s+1},\ldots,-c_n,c_p)^\top \text{ with } c_j \in [0,1] \text{ for } s+1 \le j \le p.
\end{equation}
Then, in the $(k+1)$-th iteration step, the boosting algorithm produces a vector of the form
\begin{equation}\label{beta-kplus1}
\bs{\beta}^{[k+1]} = (\underbrace{0,\ldots,0}_{\text{$s$ \textnormal{times}}}, -\tilde{c}_{s+1},\ldots,-\tilde{c}_n,\tilde{c}_p)^\top \text{ with } \tilde{c}_j \in [0,1] \text{ for } s+1 \le j \le p.
\end{equation}
\end{lemma} 
Since the vector $\bs{\beta}^{[0]} = \bs{0} \in \reals^p$ obviously has the form \eqref{beta-k}, Lemma \ref{lemma-counterexample} and a simple induction argument yield that for any $k \ge 0$, the vector $\bs{\beta}^{[k]}$ produced by the boosting algorithm is of the form $\bs{\beta}^{[k]} = (0,\ldots,0, -c_{s+1}^{[k]},\ldots,-c_n^{[k]},c_p^{[k]})^\top$ with $c_j^{[k]} \in [0,1]$ for $s+1 \le j \le p$. From this, it immediately follows that $\| \bs{\beta}^{[k]} - \bs{\beta} \|_1 \ge \| \bs{\beta}_S \|_1 = s$ for any $k \ge 0$, which in turn implies that $\| \bs{\beta}^{[k]} - \bs{\beta} \|_1 \not\rightarrow 0$ as $k \rightarrow \infty$. To complete the proof of Theorem \ref{theo-counterexample}, it remains to verify Lemma \ref{lemma-counterexample}. 
\vspace{10pt}

\textbf{Proof of Lemma \ref{lemma-counterexample}.} Since
\[ \| \bs{X}_j \|_2 = 
\begin{cases} 
\gamma                 & \text{for } 1 \le j \le s \\
1                      & \text{for } s+1 \le j \le n \\
\sqrt{(\gamma^2-1)s+n} & \text{for } j = p
\end{cases}
\]
and
\begin{align*} 
\bs{R}^{[k]} 
 & = \bs{Y} - \bs{X} \bs{\beta}^{[k]} = \bs{X} (\bs{\beta} - \bs{\beta}^{[k]}) \\
 & = (\underbrace{\gamma(1-c_p),\ldots,\gamma(1-c_p)}_{\text{$s$ times}}, c_{s+1}-c_p,\ldots,c_n-c_p)^\top, 
\end{align*}
we get that 
\[ \rho_j^{[k]} := \frac{\bs{X}_j^\top \bs{R}^{[k]}}{\| \bs{X}_j \|_2} = 
\begin{cases}
\gamma (1-c_p)  & \text{for } 1 \le j \le s \\
c_j - c_p       & \text{for } s+1 \le j \le n \\[0.2cm]
\displaystyle{\frac{s \gamma^2(1-c_p) + \sum_{j=s+1}^n(c_j-c_p)}{\sqrt{(\gamma^2-1)s+n}}} & \text{for } j = p.
\end{cases}
\]
The boosting algorithm picks the index $j_{k+1} = \arg\max_{1\le j \le p} |\rho_j^{[k]}|$ in the $(k+1)$-th iteration step. (In case of ties, $j_{k+1}$ is the smallest index $j$ with $|\rho_{j}^{[k]}| \ge |\rho_i^{[k]}|$ for all $i$.) If $\max_{1 \le j \le s} |\rho_j^{[k]}| \ge \max_{s+1 \le j \le n} |\rho_j^{[k]}|$, that is, if $\gamma(1-c_p) \ge \max_{s+1 \le j \le n} |c_j - c_p|$, then   
\begin{align*}
\rho_p^{[k]} 
 & \ge \frac{s \gamma^2(1-c_p) - (n-s) \max_{s+1 \le j \le n} |c_j - c_p|}{\sqrt{(\gamma^2-1)s+n}} \\
 & \ge \gamma(1-c_p) \underbrace{\frac{s \gamma - n + s}{\sqrt{(\gamma^2-1)s+n}}}_{> 1 \text{ for all } n \, \ge \, 4},  
\end{align*}
which implies that $|\rho_p^{[k]}| > \max_{1 \le j \le s} |\rho_j^{[k]}|$ for any $n \ge 4$. Hence, only two cases are possible: 
\begin{enumerate}[leftmargin=0.8cm,label=(\Alph*)]
\item \label{caseA} $j_{k+1} = p$, or put differently, $|\rho_p^{[k]}| > |\rho_j^{[k]}|$ for all $j \ne p$.  
\item \label{caseB} $j_{k+1} \in \{s+1,\ldots,n\}$, or put differently, there exists $j^* \in \{s+1,\ldots,n\}$ such that $|\rho_{j^*}^{[k]}| > |\rho_j^{[k]}|$ for all $1 \le j \le s$ and $|\rho_{j^*}^{[k]}| \ge |\rho_j^{[k]}|$ for all $s+1 \le j \le p$.
\end{enumerate}
In case \ref{caseA}, $\bs{\beta}^{[k+1]} = (0,\ldots,0,-\tilde{c}_{s+1},\ldots,-\tilde{c}_n, \tilde{c}_p)^\top$, where $\tilde{c}_j = c_j$ for $s+1 \le j \le n$ and $\tilde{c}_p = c_p + \nu \Delta$ with 
\[ \Delta = \frac{s \gamma^2(1-c_p) + \sum_{j=s+1}^n(c_j-c_p)}{(\gamma^2-1)s+n}. \]
The parameter $\tilde{c}_p = c_p + \nu \Delta$ has the following properties: 
\begin{enumerate}[label=(\roman*),leftmargin=0.8cm]
\item \label{prop1} Since 
\[ \Delta \le \frac{s \gamma^2(1-c_p) + (n-s)}{(\gamma^2-1)s+n} = 1 - c_p \frac{s \gamma^2}{s \gamma^2 + n - s} \le 1 - c_p, \]
it holds that $\tilde{c}_p \le c_p + \nu (1-c_p) \le 1$. 
\item \label{prop2} If $\gamma(1-c_p) \ge \max_{s+1 \le j \le n} |c_j - c_p|$, then 
\[ \Delta \ge \gamma(1-c_p) \frac{s \gamma - n + s}{(\gamma^2-1)s+n} = \gamma(1-c_p) \frac{n^{3/2} - n + n^{1/2}}{n^{5/2} + n - n^{1/2}} \ge 0, \] 
which implies that $\tilde{c}_p \ge c_p \ge 0$. 
\item \label{prop3} If $\gamma(1-c_p) < \max_{s+1 \le j \le n} |c_j - c_p|$, then $1 - c_p < \gamma^{-1} = n^{-1}$ and thus $c_p > 1 - n^{-1}$. Noticing that $\Delta \ge -(n-s)/([\gamma^2-1]s+n)$, we obtain that for any $n \ge 2$, 
\[ \tilde{c}_p \ge 1 - \frac{1}{n} - \nu \underbrace{\frac{n-s}{(\gamma^2-1)s+n}}_{\le 1/2 \text{ for any } n \ge 2} \ge 0. \] 
\end{enumerate}
Taken together, \ref{prop1}--\ref{prop3} imply that $\tilde{c}_p \in [0,1]$. Hence, in case \ref{caseA}, $\bs{\beta}^{[k+1]}$ has the form \eqref{beta-kplus1} with $\tilde{c}_j \in [0,1]$ for all $s+1 \le j \le p$. We now turn to case \ref{caseB}. Assuming without loss of generality that $j_{k+1} = s+1$, we obtain that $\bs{\beta}^{[k+1]} = (0,\ldots,0,-\tilde{c}_{s+1},\ldots,-\tilde{c}_n, \tilde{c}_p)^\top$ with $\tilde{c}_{s+1} = (1-\nu) c_{s+1} + \nu c_p$ and $\tilde{c}_j = c_j$ for $s+2 \le j \le p$. Since $c_{s+1} \in [0,1]$ and $c_p \in [0,1]$ by assumption, it immediately follows that $\tilde{c}_{s+1} \in [0,1]$. Hence, in case \ref{caseB}, $\bs{\beta}^{[k+1]}$ has the desired form \eqref{beta-kplus1} as well with parameters $\tilde{c}_j \in [0,1]$ for all $s+1 \le j \le p$. \qed

\subsection{Proof of Corollary \ref{corollary-counterexample}}\label{subsec-result-proof-corollary}

Let $\bs{X}$ and $\bs{\beta}$ be defined as in the proof of Theorem \ref{theo-counterexample}. We make use of the following two facts: 
\begin{enumerate}[label=(\roman*),leftmargin=0.75cm]

\item According to the proof of Theorem \ref{theo-counterexample}, for any $k \ge 0$, the vector $\bs{\beta}^{[k]}$ has the form $\bs{\beta}^{[k]} = (0,\ldots,0,-c_{s+1}^{[k]},\ldots,-c_n^{[k]},c_p^{[k]})^\top$ with $c_j^{[k]} \in [0,1]$ for $s+1 \le j \le p$, which implies that $\Delta^{[k]} = \bs{\beta}^{[k]} - \bs{\beta} = (-1,\ldots,-1,-c_{s+1}^{[k]},\ldots,-c_n^{[k]},c_p^{[k]})^\top$. 

\item Since $\| \bs{Y} - \bs{X} \bs{\beta}^{[k]} \|_2 = \| \bs{X} \Delta^{[k]} \|_2 \rightarrow 0$ as $k \rightarrow \infty$ by Theorem 5.1 in Temlyakov \cite{Temlyakov2000}, the vector $\Delta^{[k]}$ must converge to an element of the nullspace $\mathcal{N}(\bs{X})$ as $k \rightarrow \infty$, in particular to the vector $\Delta^* = (-1,\ldots,-1,1)^\top \in \reals^p$. 

\end{enumerate}
Taken together, (i) and (ii) yield that $\| \Delta_S^{[k]} \|_1 = s = \sqrt{n}$ for any $k \ge 0$ and $\| \Delta_{S^c}^{[k]} \|_1 \ge (p-s)/2 = (n+1-\sqrt{n})/2$ for $k$ large enough. Consequently, 
\[ \frac{n+1-\sqrt{n}}{2\sqrt{n}} \| \Delta_S^{[k]} \|_1 \le \| \Delta_{S^c}^{[k]} \|_1 \]
for sufficiently large $k$, or put differently, $\Delta^{[k]} \notin \mathbb{C}(S,\cst)$ for any $\cst < (n+1-\sqrt{n})/(2\sqrt{n})$ and sufficiently large $k$. From this, the statement of Corollary \ref{corollary-counterexample} easily follows.

\bibliographystyle{ims}
{\small
\setlength{\bibsep}{0.55em}
\bibliography{bibliography}}

\begin{thebibliography}{23}
\expandafter\ifx\csname natexlab\endcsname\relax\def\natexlab#1{#1}\fi
\expandafter\ifx\csname url\endcsname\relax
  \def\url#1{\texttt{#1}}\fi
\expandafter\ifx\csname urlprefix\endcsname\relax\def\urlprefix{URL }\fi
\providecommand{\eprint}[2][]{\url{#2}}

\bibitem[{Bickel et~al.(2009)Bickel, Ritov and
  Tsybakov}]{BickelRitovTsybakov2009}
\textsc{Bickel, P.~J.}, \textsc{Ritov, Y.} and \textsc{Tsybakov, A.} (2009).
\newblock Simultaneous analysis of {L}asso and {D}antzig selector.
\newblock \textit{Ann. Statist.}, \textbf{37} 1705--1732.

\bibitem[{B{\"u}hlmann(2006)}]{Buehlmann2006}
\textsc{B{\"u}hlmann, P.} (2006).
\newblock Boosting for high-dimensional linear models.
\newblock \textit{Ann. Statist.}, \textbf{34} 559--583.

\bibitem[{B{\"u}hlmann and Yu(2006)}]{BuehlmannYu2006}
\textsc{B{\"u}hlmann, P.} and \textsc{Yu, B.} (2006).
\newblock Sparse boosting.
\newblock \textit{Journal of Machine Learning Research}, \textbf{7} 1001--1024.

\bibitem[{Cand{\`e}s and Tao(2005)}]{CandesTao2005}
\textsc{Cand{\`e}s, E.} and \textsc{Tao, T.} (2005).
\newblock Decoding by linear programming.
\newblock \textit{IEEE Trans. Info Theory}, \textbf{51} 4203--4215.

\bibitem[{Cand{\`e}s and Tao(2007)}]{CandesTao2007}
\textsc{Cand{\`e}s, E.} and \textsc{Tao, T.} (2007).
\newblock The {D}antzig selector: statistical estimation when $p$ is much
  larger than $n$.
\newblock \textit{Ann. Statist.}, \textbf{35} 2313--2351.

\bibitem[{Chen et~al.(1998)Chen, Donoho and Saunders}]{ChenDonohoSaunders1998}
\textsc{Chen, S.}, \textsc{Donoho, D.} and \textsc{Saunders, M.} (1998).
\newblock Atomic decomposition by basis pursuit.
\newblock \textit{SIAM J. Sci. Comput.}, \textbf{20} 33--61.

\bibitem[{Cohen et~al.(2009)Cohen, Dahmen and DeVore}]{CohenDahmenDeVore2009}
\textsc{Cohen, A.}, \textsc{Dahmen, W.} and \textsc{DeVore, R.~A.} (2009).
\newblock Compressed sensing and best $k$-term approximation.
\newblock \textit{Journal of the American Mathematical Society}, \textbf{22}
  211--231.

\bibitem[{Donoho and Elad(2003)}]{DonohoElad2003}
\textsc{Donoho, D.} and \textsc{Elad, M.} (2003).
\newblock Maximal sparsity representation via $\ell_1$ minimization.
\newblock \textit{Proc. Natl. Acad. Sci.}, \textbf{100} 2197--2202.

\bibitem[{Donoho et~al.(2006)Donoho, Elad and
  Temlyakov}]{DonohoEladTemlyakov2006}
\textsc{Donoho, D.}, \textsc{Elad, M.} and \textsc{Temlyakov, V.} (2006).
\newblock Stable recovery of sparse overcomplete representations in the
  presence of noise.
\newblock \textit{IEEE Trans. Info Theory}, \textbf{52} 6--18.

\bibitem[{Donoho and Huo(2001)}]{DonohoHuo2001}
\textsc{Donoho, D.} and \textsc{Huo, X.} (2001).
\newblock Uncertainty principles and ideal atomic decomposition.
\newblock \textit{IEEE Trans. Info Theory}, \textbf{47} 2845--2862.

\bibitem[{Efron et~al.(2004)Efron, Hastie, Johnstone and
  Tibshirani}]{Efronetal2004}
\textsc{Efron, B.}, \textsc{Hastie, T.}, \textsc{Johnstone, I.} and
  \textsc{Tibshirani, R.} (2004).
\newblock Least angle regression.
\newblock \textit{Ann. Statist.}, \textbf{32} 407--499.

\bibitem[{Feuer and Nemirovski(2003)}]{FeuerNemirovski2003}
\textsc{Feuer, A.} and \textsc{Nemirovski, A.} (2003).
\newblock On sparse representation in pairs of bases.
\newblock \textit{IEEE Trans. Info Theory}, \textbf{49} 1579--1581.

\bibitem[{Freund and Schapire(1996)}]{FreundSchapire1996}
\textsc{Freund, Y.} and \textsc{Schapire, R.} (1996).
\newblock Experiments with a new boosting algorithm.
\newblock In \textit{Proceedings of the Thirteenth International Conference on
  Machine Learning}. Morgan Kaufmann, San Francisco, 148--156.

\bibitem[{Friedman(2001)}]{Friedman2001}
\textsc{Friedman, J.} (2001).
\newblock Greedy function approximation: a gradient boosting machine.
\newblock \textit{Ann. Statist.}, \textbf{29} 1189--1232.

\bibitem[{Gribonval and Nielsen(2003)}]{GribonvalNielsen2003}
\textsc{Gribonval, R.} and \textsc{Nielsen, M.} (2003).
\newblock Sparse representations in unions of bases.
\newblock \textit{IEEE Trans. Inform. Theory}, \textbf{49} 3320--3325.

\bibitem[{Hepp et~al.(2016)Hepp, Schmid, Gefeller, Waldmann and
  Mayr}]{Heppetal2016}
\textsc{Hepp, T.}, \textsc{Schmid, M.}, \textsc{Gefeller, O.},
  \textsc{Waldmann, E.} and \textsc{Mayr, A.} (2016).
\newblock Approaches to regularized regression -- a comparison between
  {G}radient {B}oosting and the {L}asso.
\newblock \textit{Methods Inf Med}, \textbf{55} 422--430.

\bibitem[{Mallat and Zhang(1993)}]{MallatZhang1993}
\textsc{Mallat, S.} and \textsc{Zhang, Z.} (1993).
\newblock Matching pursuits with time-frequency dictionaries.
\newblock \textit{IEEE Trans. Signal Process.}, \textbf{41} 3397--3415.

\bibitem[{Qian and Chen(1994)}]{QianChen1994}
\textsc{Qian, S.} and \textsc{Chen, D.} (1994).
\newblock Signal representation using adaptive normalized {G}aussian functions.
\newblock \textit{Signal Process.}, \textbf{36} 1--11.

\bibitem[{Schapire(1990)}]{Schapire1990}
\textsc{Schapire, R.} (1990).
\newblock The strength of weak learnability.
\newblock \textit{Machine Learning}, \textbf{5} 197--227.

\bibitem[{Temlyakov(2000)}]{Temlyakov2000}
\textsc{Temlyakov, V.} (2000).
\newblock Weak greedy algorithms.
\newblock \textit{Adv. Comput. Math.}, \textbf{12} 213--227.

\bibitem[{Temlyakov(2011)}]{Temlyakov2011}
\textsc{Temlyakov, V.} (2011).
\newblock \textit{Greedy approximation}, vol.~20 of \textit{Cambridge
  Monographs on Applied and Computational Mathematics}.
\newblock Cambridge University Press.

\bibitem[{Tibshirani(1996)}]{Tibshirani1996}
\textsc{Tibshirani, R.} (1996).
\newblock Regression shrinkage and selection via the lasso.
\newblock \textit{J. Roy. Statist. Soc. Ser. B}, \textbf{58} 267--288.

\bibitem[{Tropp(2004)}]{Tropp2004}
\textsc{Tropp, J.} (2004).
\newblock Greed is good: algorithmic results for sparse approximation.
\newblock \textit{IEEE Trans. Info Theory}, \textbf{50} 2231--2242.

\end{thebibliography}

\end{document}